# FROM GENES TO TOKENS: A GWAS-INSPIRED APPROACH FOR INTERPRETABLE STYLOMETRIC ANALYSIS

(This is a preprint version of a paper accepted for presentation at DH2026 and for publication in the conference proceedings)

**Dmitry Pronin, Evgeny Kazartsev**

HSE University

The study was implemented in the framework of the Basic Research Program at HSE University (HSE-BR-2025-031) in 2026.

## Abstract

This short paper introduces a stylometric interpretation method inspired by genome-wide association studies (GWAS). Each "gene" token's association with "phenotype" authorship is tested using logistic regression with multiple-comparison correction. Applied to English, German, and Russian corpora, the method detects statistically significant lexical markers distinctive of individual authors.

## 1. Introduction

Computational stylometry is a field of the Digital Humanities dedicated to authorship attribution through the identification and quantitative analysis of stylistic features. This methodology enables the identification of an author, whether unknown or disputed, by relying on statistically significant linguistic regularities. The central theoretical assumption of stylometric analysis is that a person's linguistic behavior displays stable unconscious patterns, which result in the stylistic consistency of texts produced by the same author. Although modern deep learning methods (such as transformer-based models) provide high accuracy, they often sacrifice interpretability for performance (Huang et al. 2025)., whereas interpretability has remained a central concern in stylometry from classical authorship-attribution studies to recent work on characteristic features and explainable Delta-style measures (Stamatatos 2009; Burrows 2002; Evert et al. 2017; Klaussner et al. 2015; Savoy 2012).

This study proposes an approach that applies the principles of genome-wide association studies (GWAS) to interpret differences between authors at the level of words or n-grams. In classical GWAS, each single nucleotide polymorphism (SNP) is tested for association with a specific phenotype (trait), such as a disease or a quantitative characteristic like body weight. The GWAS methodology involves fitting a separate regression model for each SNP, estimating its effect size and p-value, followed by control of false positives through significance-threshold correction for multiple testing (Uffelmann et al. 2021; Visscher et al. 2017). This makes it possible to move from a single aggregate distance between texts to a second step of interpretation: a feature-wise map of differences between already known authors.

## 2. Stylometric "GWAS"

Mathematically an approach analogous to GWAS can be applied to stylometric text features. Standardized token frequencies play the role of «genes», while the «phenotype» corresponds to the authorship of a text. Classical GWAS is typically used for binary (case-control) or quantitative phenotypes; however, the proposed framework can be extended to multiclass problems, for instance by decomposing them into a set of one-vs-rest binary tasks. Unlike Burrows's Delta, which aggregates standardized deviations across many frequent words into a single distance (Burrows 2002; Evert et al. 2017), the present framework keeps inference at the level of individual tokens and therefore produces directly interpretable lexical effect estimates.

In the case of binary attribution, logistic regression is a natural choice: the binary authorship indicator serves as the target variable. For each token τ, a separate univariate model of the form logit $P(y=1)=\alpha+\beta_\tau z_\tau$ is fitted, where $z_\tau$ is the standardized frequency of the token in a chunk. The coefficient $\beta_\tau$ estimates the direction and strength of association with authorship, while the corresponding p-value tests the null hypothesis $H_0:\beta_\tau=0$. This procedure is closer to feature-wise significance testing than to PCA loadings or other latent-space summaries, because each token receives its own signed effect size on a common inferential scale.

Since the number of analyzed features can reach thousands, the significance threshold must be adjusted to minimize the number of false positives. In GWAS, this is typically addressed by controlling either the family-wise error rate or the false discovery rate. The simplest and most common approach is the Bonferroni correction, in which the nominal significance level is divided by the number of tests. In this sense, the method is complementary to Delta-, Zeta-, or specific-vocabulary approaches: rather than replacing corpus-level stylistic comparison, it adds a transparent inferential layer for ranking candidate marker words (Figure 1).

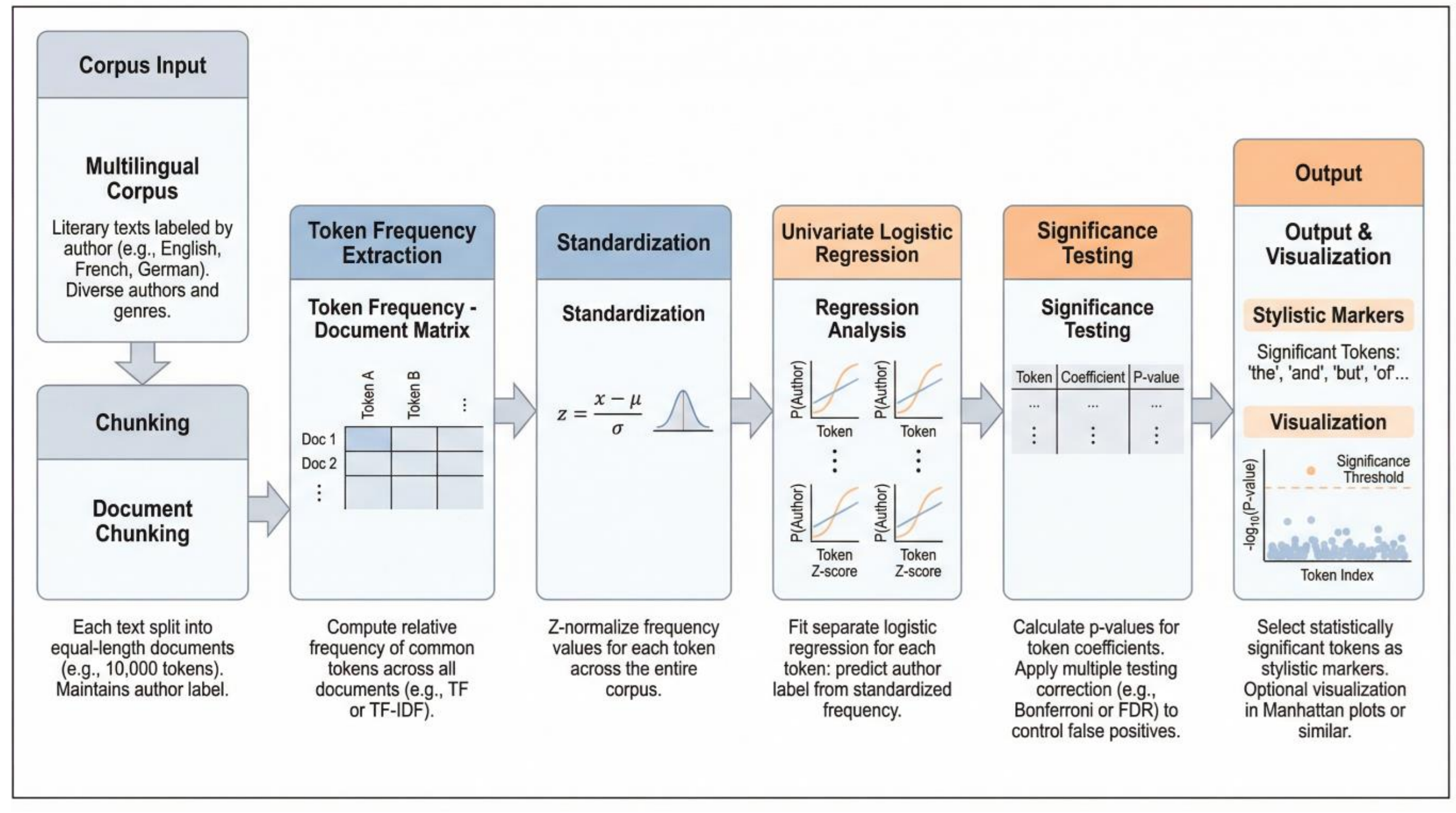

**Figure 1**. Stylometric "GWAS" pipeline.

## 3. Experiments and results

The GWAS-stylometry method was applied to texts by three authors across three corpora English, German, and Russian. Each corpus contains 75 documents with three texts per each author. Unigrams were used as tokens. The corpus was lemmatized and each work was divided into non-overlapping chunks of 10,000 lemmas. The 5,000 most frequent lemmas in each corpus were selected for analysis. After computing p-values for every lemmatized unigram Bonferroni correction was applied.

*H. G. Wells.*

The method identified 221 significant unigrams associated with the author’s distinctive style (Figure 2). The most prominent markers combine high-frequency function words with semantically meaningful items (e.g., *empire*, *tremendous*, *anyhow*), reflecting a stylistic profile that blends narrative scale, evaluative intensity, and conversational modulation.

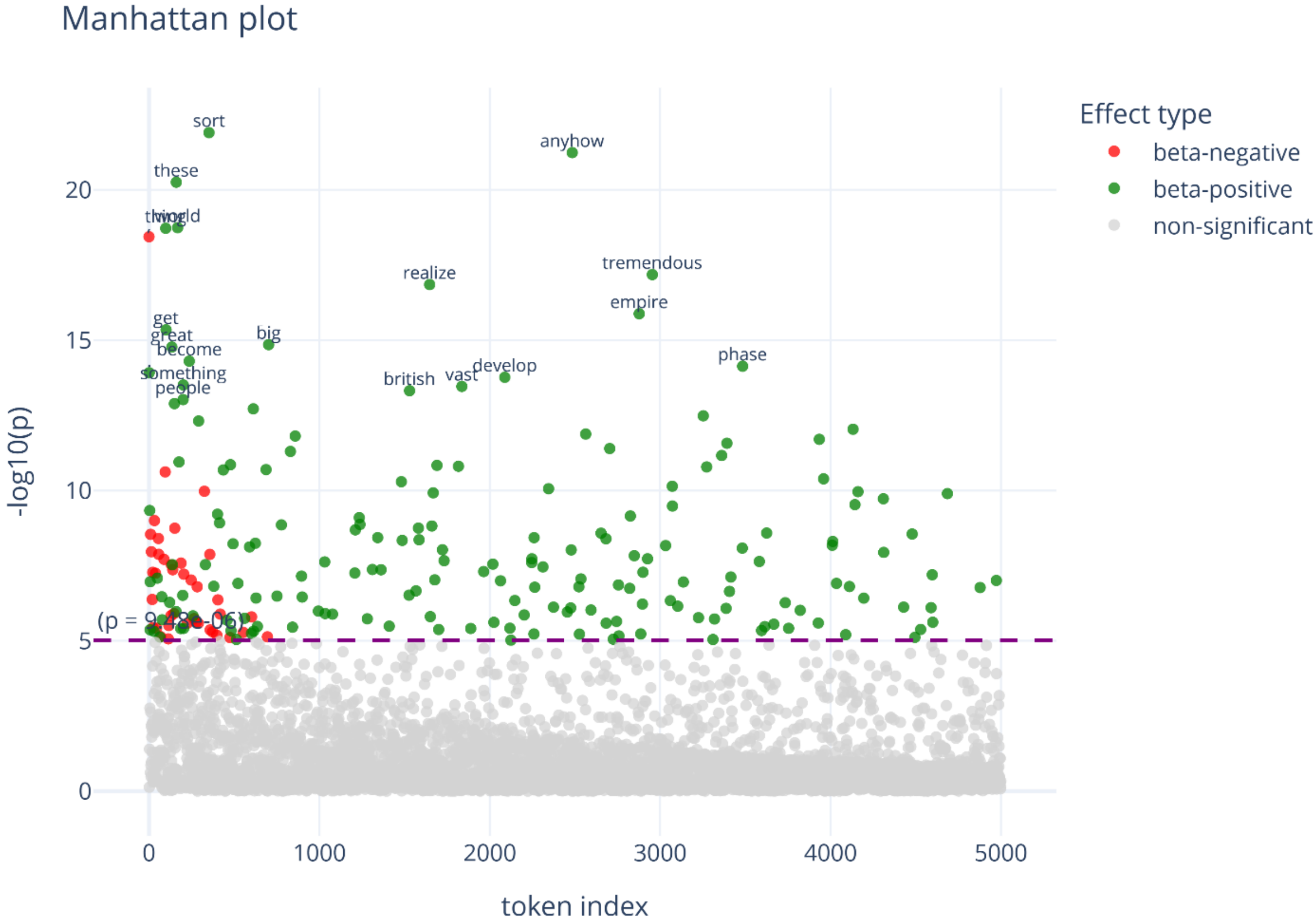

**Figure 2.** Manhattan plot for H. G. Wells. Green points represent unigrams with positive regression coefficients while red points indicate unigrams with negative coefficients. Green unigrams are common for the author whereas red ones are notably avoided in his works.

*H. Hesse.*

In the German corpus, a set of statistically significant lemmas was identified as highly specific to Hermann Hesse’s works (Figure 3). The detected markers emphasize the author’s introspective and philosophical tone. Words such as *angst* ("fear"), *erinnerung* ("memory"), *wahnsinnig* ("mad"), and *hinweg* ("away, beyond") reflect his focus on inner experience, self-reflection, and emotional intensity. Others, including *spiegeln* ("to reflect"), *still* ("quiet"), and *roch* ("smelled"), point to a nuanced attention to perception, introspection, and subtle existential states.

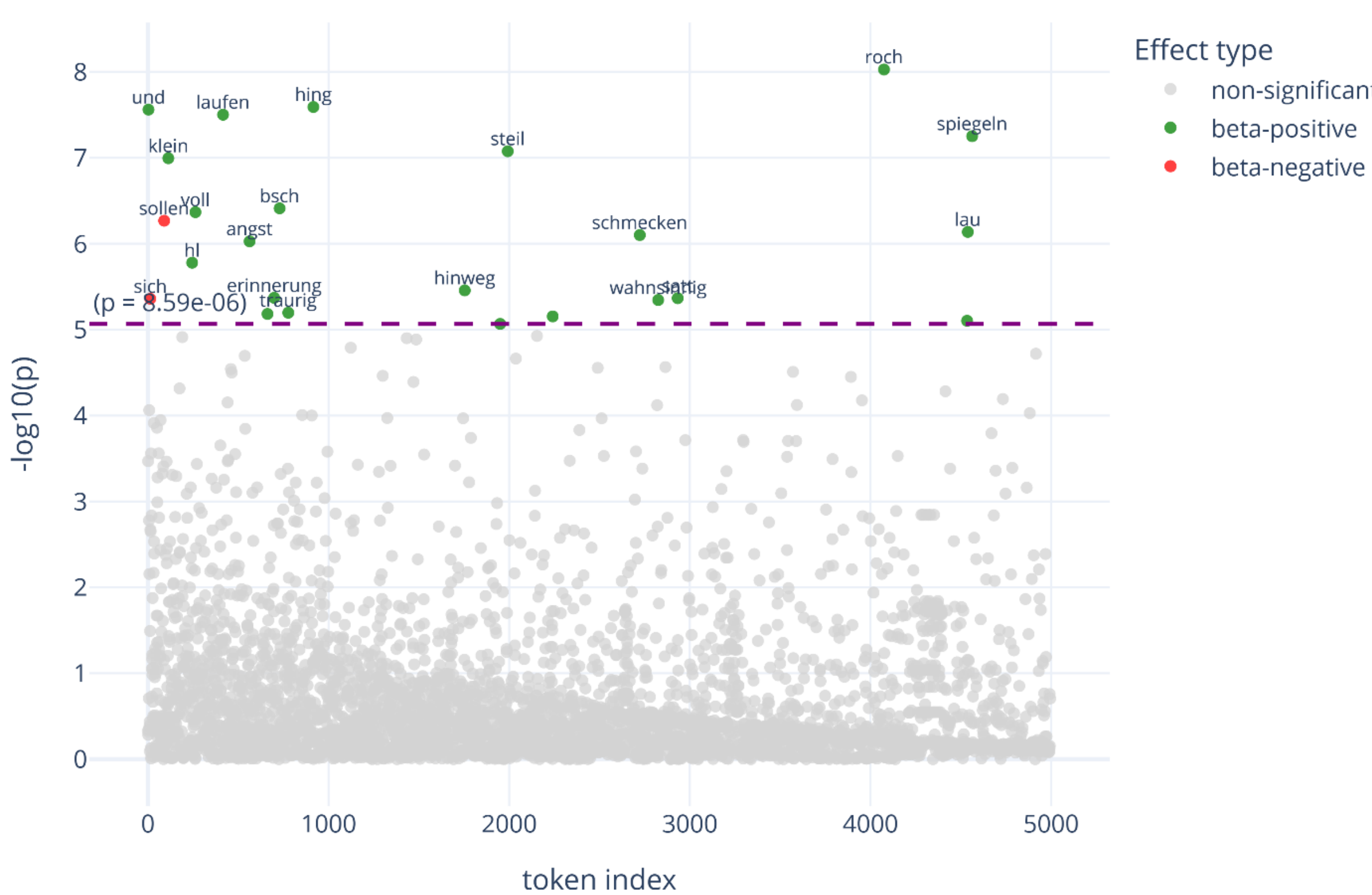

**Figure 3.** Manhattan plot for H. Hesse.

*L. N. Tolstoy.*

The analysis of Russian-language texts revealed a large set of statistically significant lemmas associated with Tolstoy's style, including both positively and negatively weighted features (Figure 4). Prominent markers such as *сказать* ("to say"), *чувствовать* ("to feel"), *увидать* ("to see"), and *услышать* ("to hear") point to a narrative mode grounded in perception and reporting. Other frequent items, including *самый* ("the most"), *очевидный* ("obvious"), and *несмотря* ("despite"), reflect patterns of emphasis, evaluation, and syntactic structuring. Taken together, these features highlight Tolstoy's realist style, characterized by detailed observation, discursive narration, and a strong orientation toward experiential and epistemic framing.

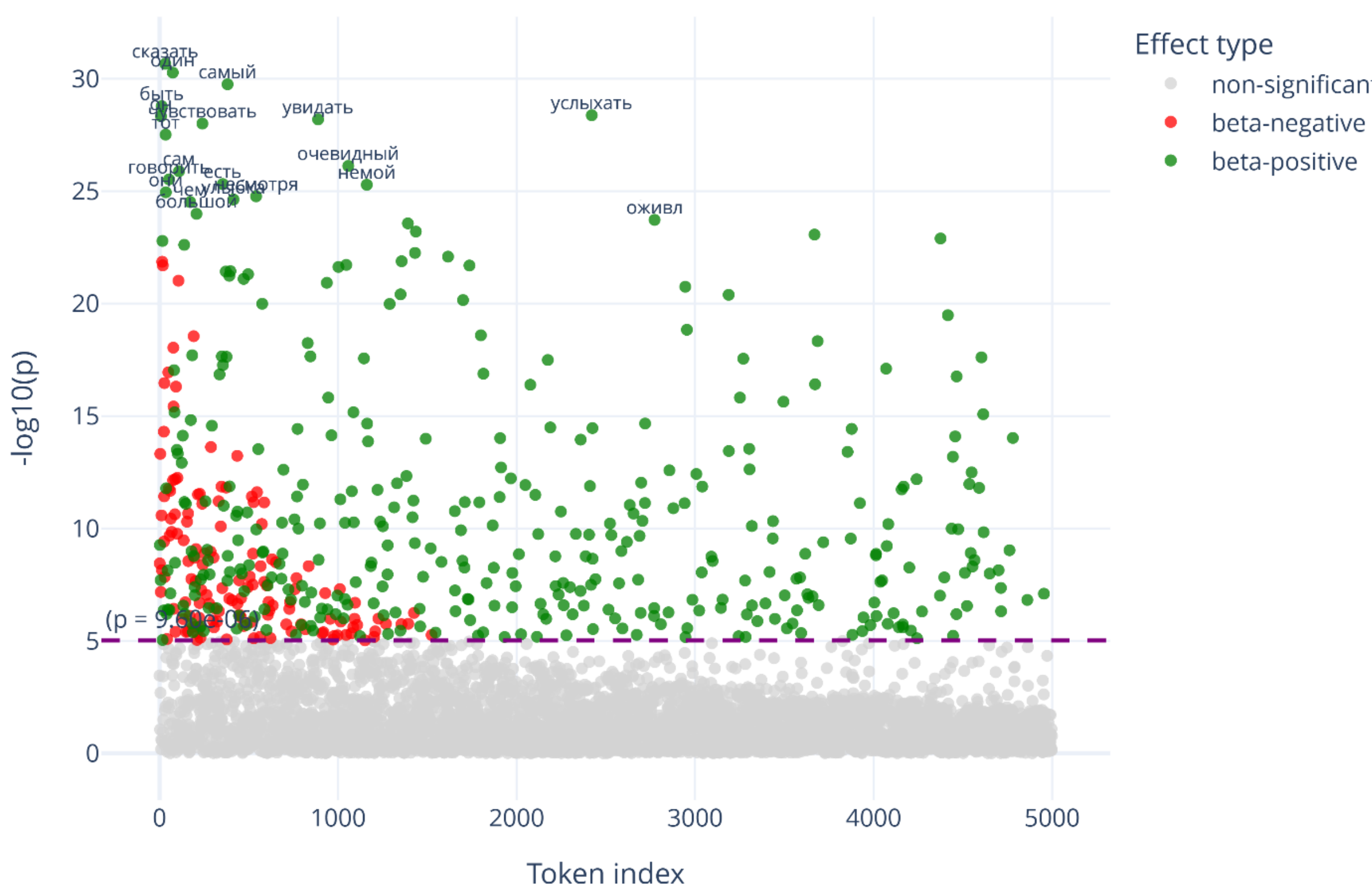

**Figure 4.** Manhattan plot for Leo Tolstoy.

The permutation test conducted for all authors demonstrated that the observed number of significant features substantially exceeded the expected count under random labeling. A public code repository with scripts for preprocessing, regression, multiple-testing correction, and Manhattan-plot visualization is being prepared and can be linked in the revised conference version to strengthen reproducibility.

## 4. Future Development

In genome-wide association studies (GWAS), it is standard practice to include covariates such as age, sex, or principal components reflecting population structure into the regression model. This helps adjust for confounding factors and prevents false associations between genetic variants and phenotypic traits. A similar principle can be applied in stylometry. For interpretation of regression results it is important to account for parameters not directly related to authorship but potentially influencing the distribution of linguistic features. Such parameters may include genre, time period of composition, and other linguistic or cultural characteristics. Incorporating these covariates allows for distinguishing genuine authorial markers from linguistic features determined by external factors. At the same time, Manhattan plots provide a complementary perspective by enabling the organization of detected signals not only by frequency, as in stylometric representations in this work, but also by predefined linguistic groupings, such as parts of speech or semantic categories. A promising direction for further development involves the use of mixed models, which can simultaneously account for fixed effects (individual stylistic markers) and random effects (contextual or genre variations), thus providing a more accurate and robust representation of stylometric structure.

## Conclusion

This work introduces an interpretable approach to authorial style analysis based on the principles of GWAS. Each token is evaluated independently for its association with authorship. When tested on different corpora (English, German, and Russian), the method effectively identified statistically significant lexical markers capturing the distinctive stylistic patterns of individual authors. Relative to aggregate distance-based measures, the approach offers an explicitly token-level, quantitatively rigorous framework that generalizes across languages and can be further extended by incorporating thematic or genre-related covariates.

## Data and Code Availability

Code and data used in this study are publicly available at:

https://github.com/DDPronin/GWAS-stylometry